\documentclass[]{spie}

\usepackage{float}
\usepackage{tabularx}
\usepackage{graphicx,graphics}
\usepackage{url}
\usepackage{amsmath,amsfonts,amssymb,textcomp,booktabs,threeparttable}
\usepackage{amssymb}
\usepackage{hhline}
\usepackage{multirow}
\usepackage[bookmarks=true,bookmarksopen=true,breaklinks=true,colorlinks=true, linkcolor=blue,anchorcolor=blue,citecolor=blue,filecolor=blue,menucolor=blue, urlcolor=blue,bookmarks=false]{hyperref}
\usepackage{stmaryrd}
\include{boldcommands}
\usepackage{tabularx}
\usepackage{booktabs}
\usepackage{eucal,epstopdf}
\newcommand{\tss}{\textsuperscript}

\title{Classifying magnetic resonance image modalities with convolutional neural networks}

\author[a,b]{Samuel Remedios}
\author[b]{Dzung L. Pham}
\author[c]{John A. Butman}
\author[b]{Snehashis Roy}
\affil[a]{Department of Computer Science, Middle Tennessee State University}
\affil[b]{Center for Neuroscience and Regenerative Medicine, Henry Jackson Foundation}
\affil[c]{Radiology and Imaging Sciences, Clinical Center, National Institute of Health}

\begin{document}
\maketitle

\begin{abstract}
Magnetic Resonance (MR) imaging allows the acquisition of images with different contrast properties 
depending on the acquisition protocol and the magnetic properties of tissues. Many MR brain image 
processing techniques, such as tissue segmentation, require multiple MR contrasts as inputs, and 
each contrast is treated differently. Thus it is advantageous to automate the identification of 
image contrasts for various purposes, such as facilitating image processing pipelines, and managing 
and maintaining large databases via content-based image retrieval (CBIR). Most automated CBIR 
techniques focus on a two-step process: extracting features from data and classifying the image 
based on these features. We present a novel 3D deep convolutional neural 
network (CNN)-based method for MR image contrast classification. The proposed CNN automatically 
identifies the MR contrast of an input brain image volume. Specifically, we explored three 
classification problems: (1) identify $T_1$-weighted ($T_1$-w), $T_2$-weighted ($T_2$-w), and 
fluid-attenuated inversion recovery (FLAIR) contrasts, (2) identify pre vs post-contrast $T_1$, (3) 
identify pre vs post-contrast FLAIR. A total of $3418$ image volumes acquired from multiple sites 
and multiple scanners were used. To evaluate each task, the proposed model was trained on $2137$ 
images and tested on the remaining $1281$ images. Results showed that image volumes were correctly 
classified with $97.57$\% accuracy.

\end{abstract}

\section{Introduction}
\label{sec:intro}
As biomedical imaging increasingly intersects with ``big data", there is a growing 
need for automated image processing and archive management. Image databases can store thousands of 
images, and assigning humans to annotate, sort, organize, and maintain every image is laborious and 
error-prone.  Additionally, different hospitals and medical scanners have their own file naming 
conventions, resulting in ambiguity and heterogeneity among medical images. Therefore
it is advantageous to automatically identify or semantically categorize images in a large database 
to facilitate searching and sorting purposes. This problem is generally central to image retrieval (CBIR) 
\cite{chang1999}, which describes automatic retrieval of images from some database based on the 
content of the image data rather than associated meta-data. For Magnetic Resonance (MR) images in 
particular, differing acquisition protocols during a scan result in different image contrast 
properties. Common MR image contrasts are $T_1$-w, $T_2$-w, $PD$-w, FLAIR, etc. The ability to automatically 
distinguish between these contrasts allows large image archives from multiple sites and scanners to 
be organized into broad categories for efficient retrieval, especially when image meta-data can
be widely inconsistent between sites and scanners. Furthermore, proper contrast identification is 
often a requirement in multi-contrast image processing pipelines for defining parameters and 
associating the appropriate training data.

CBIR involves the use of image features (or contents), such as histograms, edges, corners, blobs, 
ridges, etc., in order to categorize the desired image.  This opposes text-based image retrieval 
(TBIR), which utilizes text entries such as patient reports or human-entered meta-data. The main 
disadvantage of TBIR is that acquiring and entering these text entries can be time-consuming, 
laborious, inconsistent between sites, and prone to human error. In contrast, most CBIR techniques 
involve extraction of image features followed by their classification via machine learning, which is 
done automatically. As mentioned previously, commonly used features include edges and textures 
\cite{florea2006b}, which can be estimated and classified using k-nearest neighbors to obtain a 
semantic classification. Similarly, for multiple modalities such as computed tomography (CT), ultrasound, and MR,  
edges and local intensity distributions within a patch in the images can be used to generate a sparse 
feature dictionary \cite{srinivas2016a} of modalities. Features from a new unobserved image are 
sparsely matched to the dictionary to find most likely modality. Support vector machine (SVM)-based 
classification on wavelet features\cite{badie2013,anwar2017} and probabilistic neural networks \cite{matthew2013},  have also been previously applied to medical image CBIR and used to distinguish between normal and pathological MR brain images.

\begin{figure*}[t] 
\begin{center}
\includegraphics[width=1\textwidth]{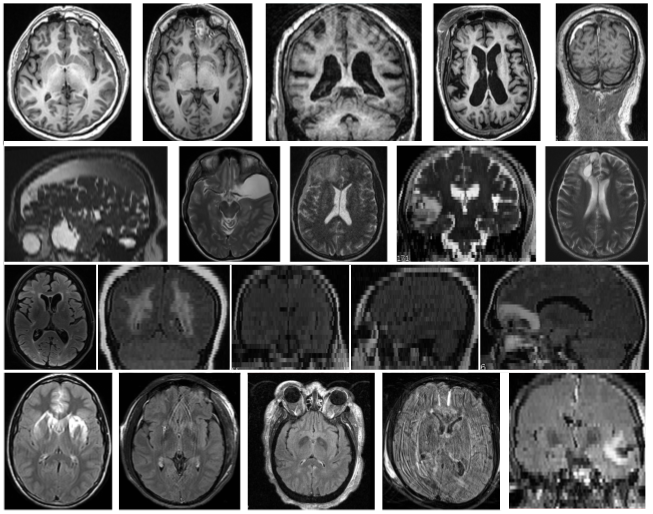}
\end{center}
\caption{Each row shows examples of pre-contrast $T_1$, $T_2$, pre and post-contrast FLAIR images, respectively, 
from our test dataset. Note the heterogeneity among brains regarding anatomy, disease 
type, resolution, field of view, and imaging artifacts such as noise, motion, ghosting, and 
intensity heterogeneity.
}
\label{fig:Fig1}
\end{figure*}

In recent years, deep learning \cite{hinton2015} and convolutional neural networks (CNN) have become more prevalent as a means to address image classification. Compared to the traditional machine learning 
classification methods such as SVM or k-nearest neighbors, CNNs do not need hand-crafted features. 
Instead, they learn and generate the necessary customized features based on the labeled training 
data provided in order to correctly classify images. CNNs have been previously used to classify 
various body parts \cite{srinivas2016b} in x-ray images. Recently, another approach used CNNs to 
classify between CT and MR images of various organs \cite{landman2015}.

The efficacy of a CNN's performance with regards to its respective task is heavily determined by its 
architecture, or the underlying structure through which the image signal passes. The prevention of 
overfitting \cite{caruana2000} and the need for regularization are commonly considered when 
designing a model with many parameters. Much work has been done to explore effective CNN 
architectures for image classification tasks in computer vision, involving state-of-the-art 
architectures such as AlexNet \cite{krizhevsky2012} in 2012, the Inception Module \cite{szegedy2014} 
in 2014, and ResNet \cite{he2016} in 2015. Each of these architectures achieved state of the art 
performance the year of their release.

In this paper, we propose a new CNN architecture called PhiNet ($\Phi$-Net), which borrows the powerful skip connection concept from the deep
residual network ResNet \cite{he2016}. $\Phi$-Net was designed with the specific purpose of classifying different contrasts 
of MR images while being robust to different types of pathologies, such as Alzheimer's disease, 
multiple sclerosis, and traumatic brain injury. Example images from our training and test dataset 
are shown in Fig.~\ref{fig:Fig1}. Although there are numerous other MR contrasts, we chose these 
broad categories because these image types are the ones most widely used in various clinical image 
processing algorithms such as tissue segmentation and registration.

Automated categorization of MR images into these broad categories helps in the homogenization of a 
``big data" imaging archive, which may contain image data from multiple sites and scanners, with or 
without visible pathologies, and where even pulse sequence names can be inconsistent. More 
specifically, we propose three tasks: 
\begin{enumerate}
\item classifying a brain MR image volume as one of $T_1$-w, $T_2$-w, or FLAIR,
\item classifying a $T_1$-w image as either pre-contrast or post-contrast,
\item classifying a FLAIR image as pre-contrast or post-contrast. 
\end{enumerate}

\begin{table}[tb]
\caption{Distribution of number of images for the three proposed tasks.
}
\label{tab:data}
\begin{center}
\begin{tabular}{lccc}
\toprule[2pt]
Classification Task & Image classes & \# Training Images & \# Testing Images\\
\toprule[1pt]
$T_1$ vs $T_2$ vs FLAIR & $T_1$ & $186$ & $227$\\ 
& $T_2$ & $189$ & $119$\\
& FLAIR & $162$ & $63$\\
\cmidrule[1pt]{3-4}
& & Total: $\mathbf{537}$ 	& Total: $\mathbf{409}$\\ 
\toprule[1pt]
$T_1$ pre vs post & $T_1$ pre	& $400$ & $204$\\
& $T_1$ post & $400$ & $374$\\
\cmidrule[1pt]{3-4}
& & Total: $\mathbf{800}$ & Total: $\mathbf{578}$\\ 
\toprule[1pt]

FLAIR pre vs post	& FLAIR pre & $400$ & $143$\\
& FLAIR post & $400$	& $151$\\
& & Total: $\mathbf{800}$ & Total: $\mathbf{294}$\\ 
\bottomrule[2pt]
\end{tabular}
\end{center}
\vspace{-1em}
\end{table}

\section{Method}
\label{sec:method}
\subsection{Data}
In total, $3418$ image volumes with various resolutions were obtained from $4$ different sites and 
$5$ different scanners: GE 3T, GE 1.5T, Philips 3T, Siemens 3T, and Siemens $1.5$T. The distribution 
of training and test data across these three tasks is shown in Table~\ref{tab:data}. Images were 
acquired for both healthy volunteers as well as patients with traumatic brain injury, hypertension, 
multiple sclerosis, and Alzheimer's disease. For training, $2137$ images were used, while the 
remaining $1281$ images were set aside and classified into the three tasks as described above. 

For the first task, we used $537$ and $409$ images for training and testing, respectively, with 
numbers of $T_1,T_2$, and FLAIR images being $186,189,$ and $162$ for training and $227,119$, and 
$63$ for testing. The second task involved $800$ ($400$ pre-$T_1$, $400$ post-$T_1$) training and 
$578$ ($204$ pre-$T_1$, $374$ post-$T_1$) testing images. For the third task, we also used $800$ 
($400$ pre-FLAIR, $400$ post-FLAIR) training and $294$ ($143$ pre-FLAIR, $151$ post-FLAIR) test 
images. To preprocess the images, the neck regions were first removed from each image volume using 
FSL \texttt{robustfov}. Then the images were re-sampled to $2\times 2\times 2$ mm\tss{3} to improve 
the processing speed for the neural network. Finally, each image was linearly scaled so that their respective $99$\tss{th} percentile of all intensities became unity.

\begin{figure}[tbh] 
\begin{center}
\includegraphics[width=1\textwidth]{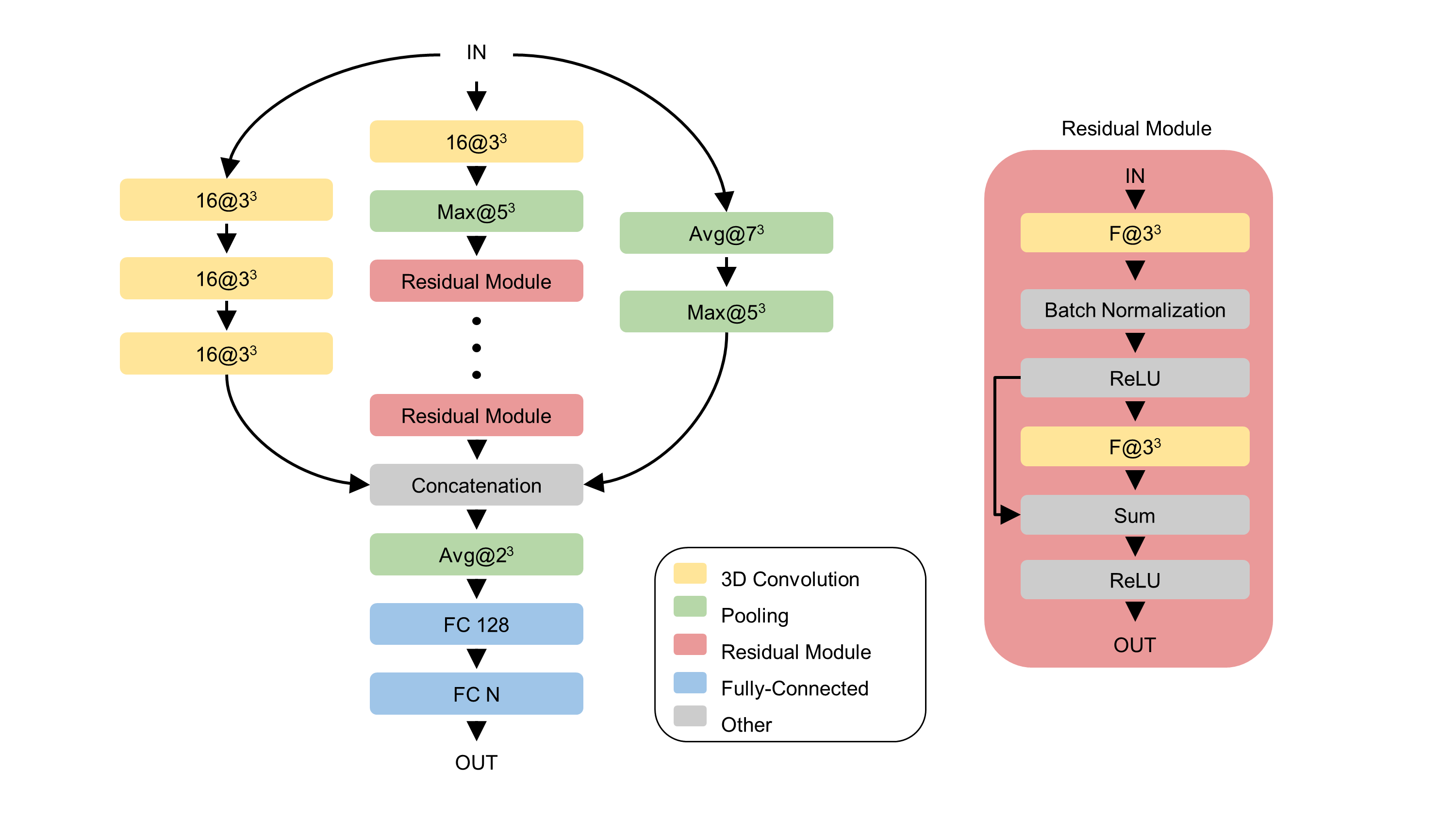}
\end{center}
\vspace{-2em}
\caption{The proposed $\Phi$-Net architecture is shown. A convolution layer, denoted by 
$N\ @\ k^3$ indicates there are $N$ filters, each having size $k\times k\times k$. 
\texttt{ReLU}, \texttt{Max} and \texttt{Avg}
denote rectified linear unit, maximum pooling, and average pooling, respectively. \textit{FC N}
denotes a fully-connected layer with N as the number of neurons. A residual module \cite{he2016}
is shown on the right, where $F$ is the number of filters for that layer in the residual module. The
$7$ layers are the same as the first $7$ layers of the original ResNet paper \cite{he2016}.
}
\label{fig:Fig2}
\end{figure}

\subsection{$\Phi$-Net 3D CNN Architecture}
Convolutional neural networks for classification purposes are usually constructed as alternating 
stacks of convolutional layers, activation layers, and pooling layers. At the end of all of the 
layers, a softmax layer is appended, producing a probabilistic prediction of the class. 
Fig.~\ref{fig:Fig2} shows the proposed $\Phi$-Net architecture. The design of this 3D CNN is to give 
three perspectives to contribute to the model's output.  The convolutional branch serves as a 
type of skip connection to preserve the original image volume signal. The residual module 
\cite{he2016} running down the center of the network allows for many high-level features to be learned, where a "high-level feature" is defined as a complex image feature such as the appearance of ventricles or cortical folds, and is composed of many low-level features such as edges, lines, and curves.  A residual module is shown 
in Fig.~\ref{fig:Fig2}. Instead of using the full $152$ residual modules as proposed in the original 
ResNet paper \cite{he2016}, we employed a smaller version of ResNet with $7$ residual 
modules. The filters used in the proposed model are the same as the first $7$ layers of the original 
ResNet paper; this was done to fit our model and 3D image data into the 4GB of GPU memory available on an NVIDIA GTX 970 GPU. The pooling branch 
preserves the overall shape but reduces the level of detail from the original image, allowing the 
model to be more robust to pathologies or other heterogeneities which could occur across an image 
population. These three stacks are then concatenated and their output is passed to a global average 
pooling layer. The output of the pooling layer is finally passed through the traditional softmax 
layer to produce the probabilities that the image belongs to each class. 

The goal with this architecture was to use these three perspectives of an incoming image in parallel 
to allow the model to learn its class, without requiring hand-crafted features such as wavelets or 
edges. The $\Phi$-Net architecture was employed with a dynamic learning rate to aid convergence. 
Categorical cross-entropy was used as the loss function for the multi-class classification task, 
i.e., $T_1$ vs $T_2$ vs FLAIR. For the binary pre vs post task, i.e. 2\tss{nd} and the 3\tss{rd} 
tasks, binary cross-entropy was chosen as the loss. For each task, the model was trained to 
convergence, defined as no decrease in validation accuracy after $100$ epochs.

As a competing method, we used a smaller version of ResNet with $11$ residual modules (called
ResNet--) in contrast to $152$ modules in the full version. This was done for speed and memory
purposes. Both neural networks were implemented in Keras \cite{chollet2015} with the Tensorflow back-end on Ubuntu 16.04 and trained with an NVIDIA GTX 970 graphics card. Note that the ResNet-- architecture lacked the convolution and the pooling branches as shown
in Fig.~\ref{fig:Fig2}. As an additional means of comparison to more traditional machine learning
methods, we also classified the test set for the first task ($T_1$-$T_2$-FLAIR comparison) via a 
registration-based method, where each test image is deform-ably registered \cite{avants2011} to a 
template $T_1$, $T_2$, and FLAIR image. Then Pearson correlation coefficients (PCC) were computed 
between each registered test image and the three templates; the template having the highest 
correlation was used as the contrast of the test image.

\begin{table}[tb]
\caption{Classification accuracy of $\Phi$-Net compared with ResNet and registration-based method.}
\label{tab:results}
\vspace{-1em}
\begin{center}
\begin{tabular}{c|cc|cc|cc}
\toprule[2pt]
& \multicolumn{2}{c|}{$T_1$ vs $T_2$ vs FLAIR} &  \multicolumn{2}{c|}{pre-$T_1$ vs post-$T_1$} &
 \multicolumn{2}{c|}{pre-FLAIR vs post-FLAIR}  \\
\hline
& ResNet-- & $\Phi$-Net & ResNet-- & $\Phi$-Net & ResNet-- & $\Phi$-Net \\
\hline
Accuracy & 98.53\% & $\mathbf{99.27\%}$ & 97.40\% & $\mathbf{99.56\%}$ & 90.48\% & $\mathbf{93.88\%}$ \\
\# Correct Predictions & 403/409  & \textbf{406/409} & 563/578 & \textbf{576/578} & 
266/294 & \textbf{276/294} \\
\# Errors & 6/409 & \textbf{3/409} & 15/578 & \textbf{2/578} & 28/294 & \textbf{18/294} \\
\bottomrule[2pt]
\end{tabular}
\end{center}
\end{table}

\begin{figure}[tb] 
\begin{center}
\includegraphics[width=1\textwidth]{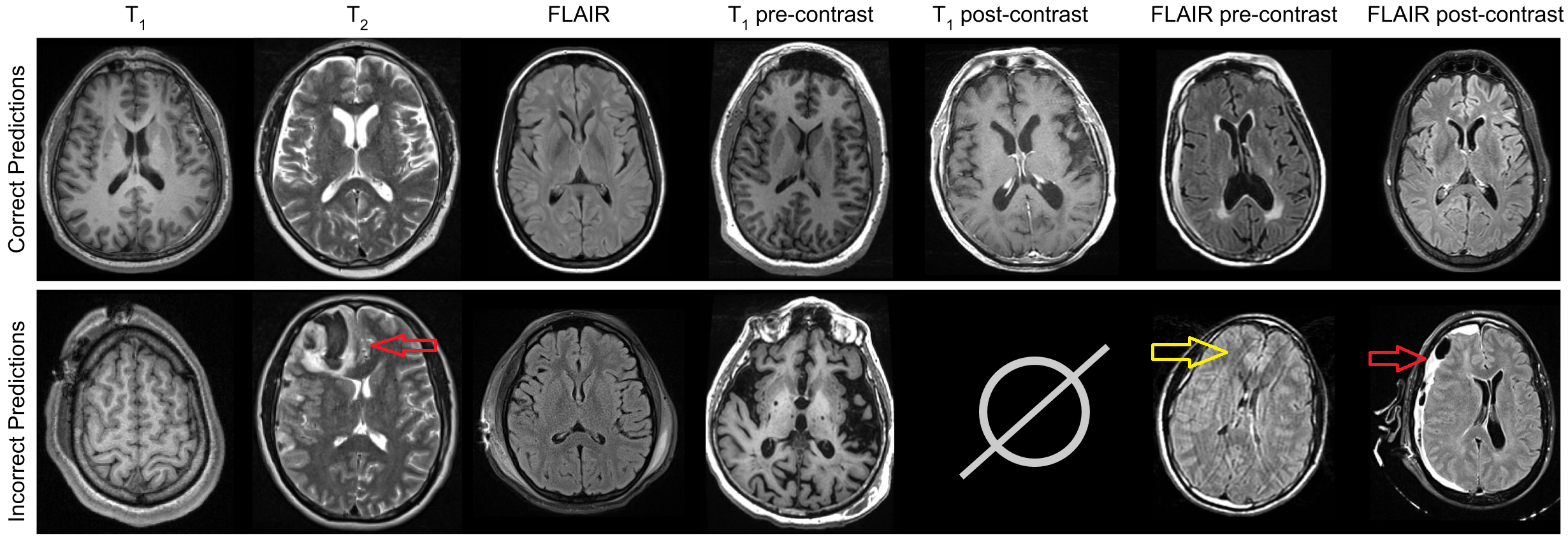}
\end{center}
\caption{Examples of correct and incorrect predictions by $\Phi$-Net are shown. First row indicates 
correct predictions, and the second row shows some errors. From left to right, the erroneous 
predictions made were: $T_2$, $T_1$, $T_1$, post-$T_1$, (no true post-$T_1$ images were 
misclassified), post-FLAIR, and post-FLAIR. Note how the majority of misclassified images have 
artifacts (yellow arrow) or severe pathologies (red arrow).
}
\label{fig:predictions}
\end{figure}

\section{Results}
Table~\ref{tab:results} shows classification accuracy of $\Phi$-Net comparing with ResNet--.
$\Phi$-Net outperforms ResNet-- in all three classification tasks. $\Phi$-Net has a mean accuracy of 
$97.57$\% over all three tasks, while ResNet-- has $95.47$\%. Although accuracies are high for both 
models, $\Phi$-Net produces  more than a $3$\% improvement for the pre-FLAIR vs post-FLAIR 
classification, which is usually the most challenging task, even for a visual comparison. To compare 
the test performances between ResNet-- and $\Phi$-Net, we performed McNemar's test \cite{mcnema1947}
over each of the three tasks, obtaining the p-values of $0.45$, $0.002$, and $0.052$ for $T_1$ vs 
$T_2$ vs FLAIR, $T_1$ pre vs post, and FLAIR pre vs post classifications, respectively. Therefore,
$\Phi$-Net produces a significantly more accurate classification between pre and post-contrast images
than ResNet--, while being similar in the case of the $T_1$-$T_2$-FLAIR classification task. Note that
while the first task is comparatively easier, pre- vs post-contrast FLAIR identification can
sometimes be difficult for a human observer, and $\Phi$-Net misclassified only $18$ of $294$ images in this
category.

Fig.~\ref{fig:predictions} shows some classification examples from the test set, with the first row 
corresponding to correct classifications made by $\Phi$-Net and the second row showing the 
incorrect ones. Of the handful of classification errors that occurred, the majority suffered from 
imaging artifacts (Fig.~\ref{fig:predictions} yellow arrow) or pathologies 
(Fig.~\ref{fig:predictions} red arrow), which confounded the model's ability to make accurate 
predictions. No post-contrast $T_1$ images were misclassified. Although both of the CNN-based
methods achieved more than $90$\% accuracy, the registration and correlation based method achieved
only an average of $81.19$\% accuracy on the $T_1$-$T_2$-FLAIR classification task. It classified
$98$\% of T$_2$ images correctly and $72$\% of both T$_1$ and FLAIR images correctly. However, its
$15$\% average lower accuracy compared to the deep learning approaches indicates that the template-based 
classification is not as robust.

\section{Discussion}
We have presented $\Phi$-Net, a novel 3D convolutional neural network architecture for the 
classification of MR brain images.  Training on an Nvidia 970 GTX took approximately $20$ hours and 
its ability to converge on differing tasks shows that $\Phi$-Net is generalizable and can be applied 
to a variety of classification problems, achieving $97.57$\% mean accuracy across $3$ tasks. Future 
work includes expanding the number of classes to categorize, one-step classification of pre vs post 
$T_1$ and FLAIR images rather than separate tasks (or alternatively, cascading the classification first as $T_1$-$T_2$-FLAIR, then as pre vs post-contrast), comparison with other CBIR techniques, 
and possible integration with a time-series model to automatically text-annotate MR images with 
human-searchable features for a smoother CBIR pipeline.

\section{ACKNOWLEDGEMENTS}
Support for this work included funding from the Intramural Research Program of the NIH and the 
Department of Defense in the Center for Neuroscience and Regenerative Medicine. This work was also 
partially supported by a grant from National MS Society RG-1507-05243.

\bibliographystyle{spiebib}
\small{
\bibliography{remedios-spie2018.bbl}
}
\end{document}